\documentclass[a4paper, 10pt, conference]{ieeeconf}      % Use this line for a4
\usepackage{FG2023}

\FGfinalcopy % *** Uncomment this line for the final submission

\IEEEoverridecommandlockouts                              % This command is only
\usepackage{graphics} % for pdf, bitmapped graphics files
\usepackage{epsfig} % for postscript graphics files
\usepackage{mathptmx} % assumes new font selection scheme installed
\usepackage{times} % assumes new font selection scheme installed
\usepackage{amsmath} % assumes amsmath package installed
\usepackage{amssymb}  % assumes amsmath package installed
\usepackage{subfigure}
\usepackage{ltxgrid}
\usepackage{capt-of}

\def\FGPaperID{****} % *** Enter the FG2023 Paper ID here

\title{\LARGE \bf
CoNFies: Controllable Neural Face Avatars
}

\author{\parbox{16cm}{\centering
    {\large Heng Yu$^1$, Koichiro Niinuma$^2$, László A. Jeni$^1$}\\
    {\normalsize
    $^1$ Carnegie Mellon University, Pittsburgh, PA, USA\\
    $^2$ Fujitsu Research of America, Pittsburgh, PA, USA}}
    \thanks{979-8-3503-4544-5/23/\$31.00 ©2023 IEEE}% <-this % stops a space
}

\begin{document}

\ifFGfinal
\thispagestyle{empty}
\pagestyle{empty}
\else
\author{Anonymous FG2023 submission\\ Paper ID 95 \\}
\pagestyle{plain}
\fi
\maketitle

\onecolumngrid
\begin{center}    
  \includegraphics[width=0.9\textwidth]{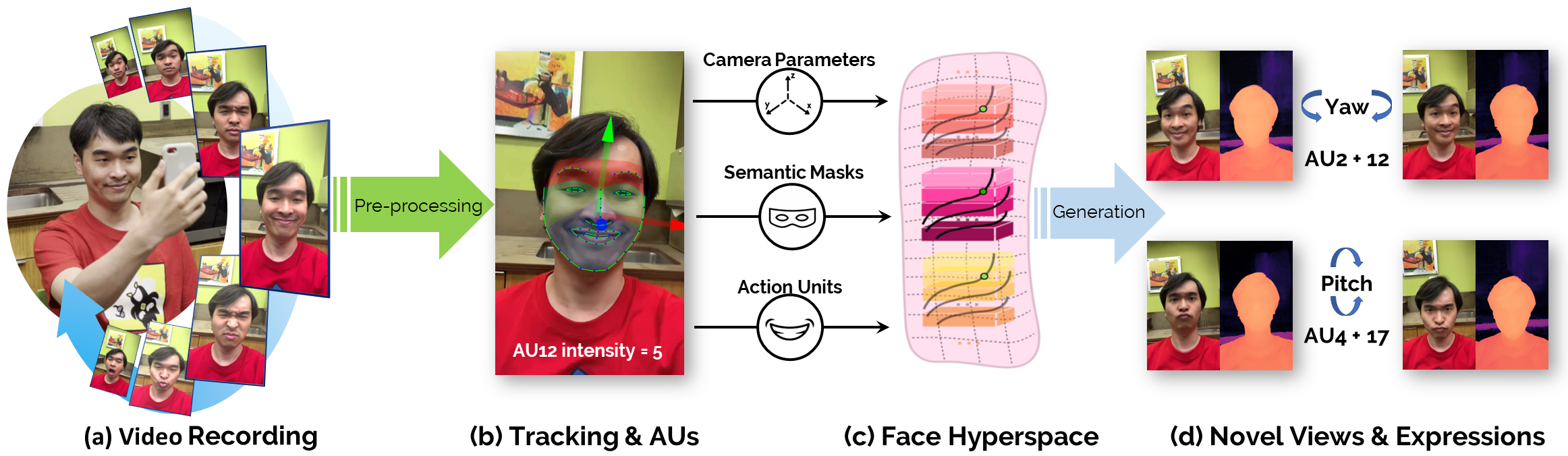}
  \captionof{figure}{2D video of users recording themselves using a circular motion during a semi-structured facial expressions task (a) is processed by a person-independent face tracker that codes facial action units (AU) (b). The estimated camera parameters, semantic masks, action unit intensities, and the original 2D frames are used to build a disentangled face hyper-space (c). From this representation, novel views and unseen expressions can be generated along with their 3D depth (d).}
  \label{pipeline}
\end{center}
\twocolumngrid

%\begin{figure*}[ht]
%  \centering
%  \includegraphics[width=0.9\textwidth]{FG2023_Latex_template/imgs/pipeline_narrow.png}
%  \caption{pipeline}
%  \label{pipeline}
%\end{figure*}

%%%%%%%%%%%%%%%%%%%%%%%%%%%%%%%%%%%%%%%%%%%%%%%%%%%%%%%%%%%%%%%%%%%%%%%%%%%%%%%%
\begin{abstract}

%While volumetric neural rendering methods like Neural radiance field (NeRF) can render photo-realistic novel-view images and some NeRF-based variants extend the neural 3D representations to non-rigid objects such as avatar, portrait attribute controlling can still be desirable and has huge potential in AR/VR applications. In this paper, we propose an automatic system that enables controlling different parts of avatars without using parametric model like 3D morphable face model (3DMM) or manually labeling attribute values or areas. To the best if our knowledge, it is the first automatic system that can be completed using only one mobile phone and no other complicated hardware required.

Neural Radiance Fields (NeRF) are compelling techniques for modeling dynamic 3D scenes from 2D image collections. These volumetric representations would be well suited for synthesizing novel facial expressions but for two problems. First, deformable NeRFs are object agnostic and model holistic movement of the scene: they can replay how the motion changes over time, but they cannot alter it in an interpretable way. Second, controllable volumetric representations typically require either time-consuming manual annotations or 3D supervision to provide semantic meaning to the scene.
We propose a controllable neural representation for face self-portraits (CoNFies), that solves both of these problems within a common framework, and it can rely on automated processing. We use automated facial action recognition (AFAR) to characterize facial expressions as a combination of action units (AU) and their intensities. AUs provide both the semantic locations and control labels for the system. CoNFies outperformed competing methods for novel view and expression synthesis in terms of visual and anatomic fidelity of expressions.

\end{abstract}

%%%%%%%%%%%%%%%%%%%%%%%%%%%%%%%%%%%%%%%%%%%%%%%%%%%%%%%%%%%%%%%%%%%%%%%%%%%%%%%%
\section{INTRODUCTION}

3D understanding of the world is crucial to the next round of technological innovations in creating digital representations of scenes, objects, and humans. However, the cost of getting 3D supervision for such systems is astronomically higher than those in 2D. Neural volumetric representations, such as Neural Radiance Fields (NeRF)~\cite{mildenhall2020nerf}, are compelling alternatives for building high fidelity representations from 2D image collections only.

%Implicit neural representation is very popular in novel view synthesis recently, among which neural radiance field (NeRF)~\cite{mildenhall2020nerf} is the one of the most important work. 

Although, previous work in this direction has demonstrated promising results for modeling and synthesizing novel views of static scenes~\cite{zhang2020nerf++}~\cite{fridovich2022plenoxels}~\cite{chen2022tensorf}~\cite{xu2022point}, articulated objects~\cite{tretschk2021non}~\cite{Pumarola20arxiv_D_NeRF}~\cite{gafni2021dynamic}, have explored the use of coarse-grain controls over limited properties, such as color~\cite{jang2021codenerf}, material~\cite{zhang2021nerfactor}, and object editing~\cite{yang2021learning}, relatively neglected is the fine-level control of semantic scene attributes. 

%In this paper, we are interested in fine-grained and comprehensive control over expressions of avatar. Some existing works 

Our interest is in general social interactions, and thus we are interested in high-fidelity 3D modeling of facial appearance and dynamics.
Previous neural representation based approaches for facial actions either required parametric models to encode facial expressions ~\cite{gafni2021dynamic}~\cite{athar2022rignerf} or were limited in the level of control over scene attributes~\cite{park2021hypernerf}. 

In a recent work Kania et al.~\cite{kania2022conerf} proposed a controllable neural representation that can achieve simple manipulation, such as opening and closing the mouth and the eyes using a learned mapping between a segmentation mask and a control value that describes the state of that region. The method relies on temporally sparse and manual annotation of the regions of interest along with their control signals, which limits the application of the method.

%The most similar work to ours is CoNeRF~\cite{kania2022conerf}, which can achieve simple control such as mouth open/close, eye open/close over avatar with the help of learned mapping between attribute value and attribute mask. CoNeRF can only synthesize simple movement regarding to each attribute and also requires manual labeling of attribute value and mask, which limits its application. 

We wish to make high fidelity 3D reconstruction and control of complex facial movements with a simplified camera setup and little or no manual annotation. A high-level summary of our method is shown on Fig.~\ref{pipeline}. First, we capture fine-scale transitions of facial movements during a semi-structured expression task. The recorded video is then processed with an automated facial action recognition system \cite{Ertugrul19_AFAR}~\cite{baltrusaitis2018openface} that provides anatomically correct action unit (AU)~\cite{rosenberg2020face} intensities and facial landmarks. From these, semantic facial masks are generated automatically and frames are sub-sampled to build an AU-balanced set of training data. The selected 2D frames, semantic masks, AU intensities, and camera parameters then used to build a face hyper-space, that can be used to synthesize novel views and unseen expression combinations.

%Thus, we propose an automatic system that can perform fine--grained and complicated control over multiple attributes of an avatar without any manual labeling. The whole pipeline is shown as Fig.~\ref{pipeline}. The training process can be summarize as: First, collect video data with a smart phone using slo-mo mode. Then perform automatic Facial Action Coding System (FACS)~\cite{rosenberg2020face} action units (AUs) and facial landmark detection on each frame. Next apply the average filter on AUs to reduce noises and also generate corresponding masks using facial landmarks, and do balanced sampling of the frames. Finally perform neural radiance field learning using the frames with AUs and mask annotations. After the system is well-trained, we can control different expressions by inputting different AU values. 

Our contributions are as follows:
%\begin{description}

%\item [\textbf{Anatomically Correct Control}] \hfill \\ Previous work was limited to holistic deformations or required manual annotation. We achieve an anatomically controllable neural avatar with no manual annotation. We show that this is achievable by using automated facial action coding that provides consistent facial keypoints and semantic labels across subjects.

%\item [\textbf{Multilabel Semantic Masks}] \hfill \\ We achieve a completely disentangled representation in the feature space where the different semantic regions do not affect each other and each region have multiple semantic control variables. We demonstrate that this formulation correctly handles different action unit combinations and achieves better visual fidelity than previous methods.

%\end{description}

\begin{itemize}
\item \textbf{Anatomically Correct Control.} Previous work was limited to holistic deformations or required manual annotation. We achieve an anatomically controllable neural avatar with no manual annotation. We show that this is achievable by using automated facial action coding that provides consistent facial key-points and semantic labels across subjects.

\item \textbf{Multi-label Semantic Masks.} We achieve a completely disentangled representation in the feature space where the different semantic regions do not affect each other and each region has multiple semantic control variables. We demonstrate that this formulation correctly handles different action unit combinations and achieves better visual fidelity than previous methods.
\end{itemize}

%\begin{itemize}
  %\item We propose an automatic avatar control system that can achieve fine-grained and comprehensive control over expressions.
  %\item We use slo-mo mode data to leverage the transition movement in the training frames.
  %\item We automatically detect the FACS AUs and key-points to generate the ground truth attribute values and masks without requiring any manual annotation.
  %\item We improve the network on the basis of CoNeRF so that each avatar region can have multiple attribute values and different regions do not affect each other.
%\end{itemize}

\section{RELATED WORKS}
Our work focuses on automatic control over avatar expressions and is closely related to several computer vision and graphics research domains such as neural rendering and avatar animation.

\subsection{Neural Rendering and Novel View Synthesis}

Implicit neural representations represented by NeRF has become more and more popular recently. NeRF can synthesize high-quality rendering results from novel views and some following extensions further enhanced the algorithm in rendering quality improvement~\cite{zhang2020nerf++}~\cite{xu2022point}, faster training and inference~\cite{garbin2021fastnerf}~\cite{yu2021plenoctrees}~\cite{muller2022instant}, generalization model~\cite{schwarz2020graf}~\cite{niemeyer2021giraffe} and so on. NeRF and its variants achieve remarkable performance on static objects, and several of these variants extend it to dynamic scenes, which is the same scenario as ours. Park et al.~\cite{park2021nerfies}~\cite{park2021hypernerf} introduce deformation fields along with a canonical NeRF to learning the movements. Some other works handle the dynamic scene through learning movement offset~\cite{pumarola2021d}~\cite{tretschk2021non} or scene flow~\cite{li2021neural}~\cite{xian2021space}. These methods achieve eye-catching results in dynamic scenes and can achieve the separation of static parts and dynamic parts to some extent through the learned deformation field/offset/flow. However, they are far from the fine-grained control over the dynamic scenes.

\subsection{Avatar Animation}
Avatar animation is a well explored research area. Some works have attempted to manipulate or edit a face~\cite{deng2020disentangled}~\cite{kowalski2020config}~\cite{tewari2020pie} while they are mainly image-based and fail to leverage 3d representation. Other works~\cite{koujan2020head2head}~\cite{kim2018deep}~\cite{tewari2020stylerig} utilize 3D Morphable Model (3DMM) as 3D face representation to achieve the head pose control and image or video reanimation. However, they have limited ability to synthesize novel views since they neglect scene geometry or appearance. Given that high-quality novel view synthesis and fine-grained avatar control is pretty challenging, some works take advantage of neural rendering to model a non-rigid 3d avatar. NerFACE~\cite{gafni2021dynamic} allows face expression and head pose control by modeling a 4D face avatar using neural radiance fields and a facial expression tracking algorithm while it assumes a static background and fixed camera. Some other works either require professional equipment and training dataset~\cite{ma2021pixel} or impose parametric face models such as 3DMM~\cite{athar2021flame}~\cite{athar2022rignerf}, which limit their application scenarios. HyperNeRF~\cite{park2021hypernerf} introduces hyper space that can better fit dynamic face avatars and also control facial expressions to some extent through hyper space. However, it is far from fine-grained control and cannot achieve per-attribute control. CoNeRF~\cite{kania2022conerf} can achieve per-attribute control by imposing an attribute value and mask based on HyperNeRF while it can only achieve simple control over each attribute and different attributes may affect each other. It also requires manual labeling which is labor-intensive. In contrast, we propose an automatic system that enables fine-grained comprehensive control over a face avatar and novel-view synthesis simultaneously without any manual labels.
Recently, Cao et al.~\cite{cao2022} proposed an approach creating volumetric avatars using only a short phone capture. Though their approach can generate a high-fidelity avatar, a large-scale high resolution multi-view dataset is required to pre-train their model. Unlike their approach, our method requires only a single input video.

\section{METHOD}
Our system consists of three parts: (i) data and annotation processing, (ii) network training, and (iii) avatar control. We will describe each component in detail in the following parts.

\subsection{Data and annotation processing}

\begin{figure}[b]
  \centering
  \includegraphics[width=0.25\textwidth]{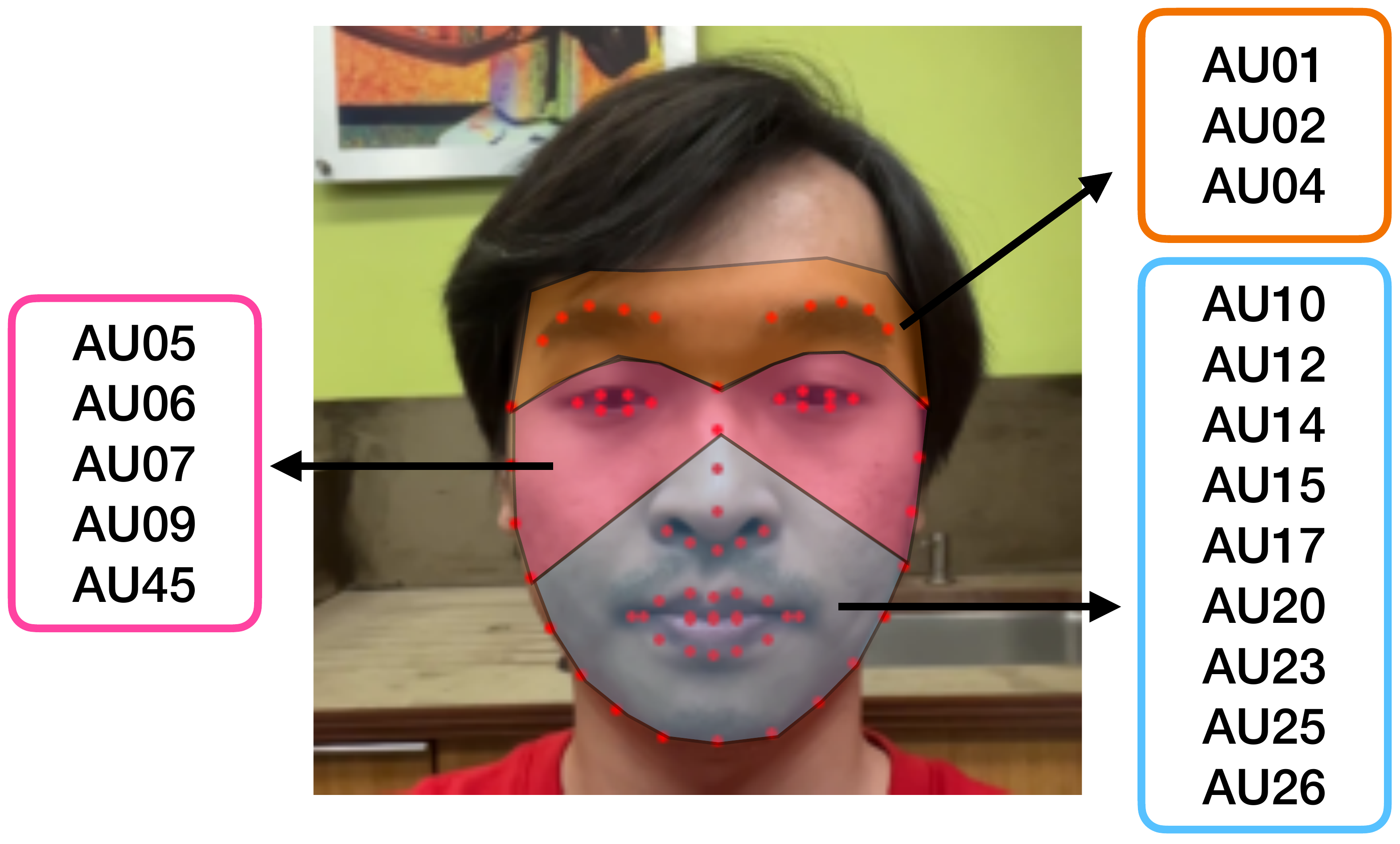}
  \caption{Facial landmarks, AUs and mask annotations}
  \label{anno}
\end{figure}

\begin{figure}
	\centering
	\subfigure[Uniform Sampling]{
		\begin{minipage}{0.20\textwidth} 
            \includegraphics[width=\textwidth]{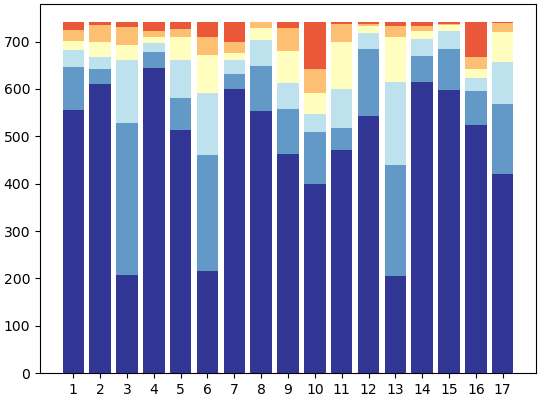} \\
		\end{minipage}
	}
	\subfigure[Balanced Sampling]{
		\begin{minipage}{0.20\textwidth}
			\includegraphics[width=\textwidth]{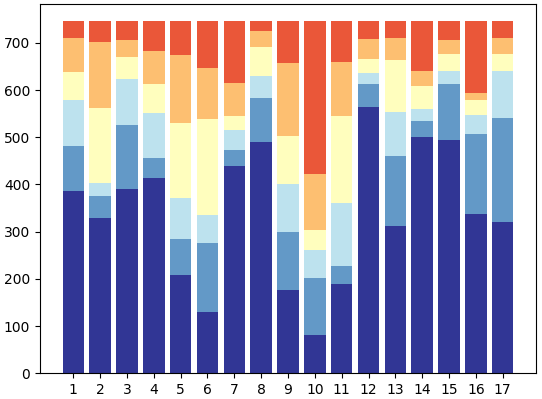} \\
			
		\end{minipage}
	}
	\caption{AU intensity distribution using different sampling strategies} 
	\label{sampling}
\end{figure}

%\subsubsection{Data collection and processing}
The data collection process of our system can be done using just a smart phone with a slo-mo video function. After collecting slo-mo video of changing expressions with moving cameras, we apply OpenFace~\cite{baltrusaitis2018openface} on every frame to detect the facial landmarks~\cite{amir2017convolutional}~\cite{baltrusaitis2013constrained} and facial action units (AUs)~\cite{baltruvsaitis2015cross}. It is worth noting that other facial landmark/AU detection methods can also work and may have better results, but this is not the focus of our work. OpenFace can output 68 2D landmark locations and 17 AU intensities from 0 to 5 (as shown in Fig.~\ref{anno}). To mitigate the noise impact of AUs detection between adjacent frames, we apply Savitzky–Golay filter~\cite{savitzky1964smoothing} to smooth the AUs and then sample the frames to reduce the computational load of the whole system. 
We found uniform sampling can lead to an extreme imbalance in AU intensity distribution (Fig.~\ref{sampling}(a)) since there exist many neutral frames in the dataset. To alleviate this problem, we propose a balanced sampling strategy. We define each AU value and AU intensity pair as a AU-intensity block, which consists of frames with a corresponding AU value and intensity. We ascendingly sort all the AU intensity blocks (total block number equals to intensity number times AU number) according to frame number in each block first. Then we uniformly sample frames from each block in order. Before sampling each block, we remove the frames that are already sampled. Using this strategy, we can get a more balanced sampling result shown in Fig.~\ref{sampling}(b).

%\subsubsection{Annotation generation }
After obtaining the 2D facial landmarks and AUs, we generate attribute masks and controllable AUs values as annotations. We define three regions and assign each action unit (AU) to its corresponding mask as shown in Fig.~\ref{anno}. We calculate the middle points of eyebrow and eye key-points as the boundary between the first and second region. We also extend some distance along the eyebrow direction as the boundary of the first region to make sure the whole eyebrows are included in the region. The boundary of the third region consists of landmarks ($\#3-\#13$, $\#28$) detected by OpenFace.
We also normalize each AU according to:
\begin{align}
\label{eq1}
  AU' = \min ( \frac{AU-AU_{min}}{\alpha AU_{max}-AU_{min}}\times 2-1, 1)
\end{align}
where $AU\in [0, 5]$, $AU'\in [-1, 1]$, $AU_{min}$ and $AU_{max}$ are minimum and maximum and values for each AU among all frames, respectively, and $\alpha$ is the factor that adjusts the maximum of AUs and we set $\alpha$ as 0.8 for all the experiments.

\subsection{Network architecture}
In this section, we briefly introduce NeRF~\cite{mildenhall2020nerf}, HyperNeRF~\cite{park2021hypernerf} and CoNeRF~\cite{kania2022conerf} for completeness and then describe our approach in detail.

\subsubsection{Neural Radiance Field (NeRF). }
NeRF uses a fully-connected neural network to learn the implicit 3D scene volumetric representations through a partial set of 2D images and can generate novel views. The NeRF network takes a sample 3d position $\textbf{x}=(x,y,z)$ and a 2d view direction $\textbf{d}=(\theta, \phi)$ as input and outputs the emitted color $\textbf{c}$ and volume density $\sigma$ at position $\textbf{x}$ with view direction $\textbf{d}$. Then one can accumulate the densities and colors into image pixels $\textbf{C}$ in RGB using classical volume rendering techniques~\cite{kajiya1984ray} as follows:
\begin{align}
\label{eq2}
  C(\textbf{r}) = \int_{t_n}^{t_f}T(t)\sigma(\textbf{r}(t))\textbf{c}(\textbf{r}(t),d)dt
\end{align}
where $T(t)=\exp{(-\int_{t_n}^t\sigma(\textbf{r}(s))ds)}$, $\textbf{r}(t)=\textbf{o}+t\textbf{d}$ is the camera ray with near bound $t_n$ and far bound $t_f$. $C(\textbf{r})$ is the expected image pixel color of the ray $\textbf{r}(t)$.

\subsubsection{HyperNeRF and CoNeRF. }
Given that original NeRF can only model static scenes, HyperNeRF, which extend Nerfies~\cite{park2021nerfies}, is proposed to model dynamic objects, especially face avatars, by introducing canonical hyper-space. The input of HyperNeRF includes sample point $\textbf{x}$ and view direction $\textbf{d}$, which is similar to the template NeRF, and also a latent deformation code $\omega_i$ and a latent appearance code $\psi_i$. The sample point $\textbf{x}$ concatenated with the image’s latent deformation code $\omega_i$ are taken as input to the spatial deformation field $T$ and the ambient slicing surface $H$ as follows, which yields a warped coordinate $\textbf{x'}$ and a coordinate in ambient space $\textbf{w}$, respectively.
\begin{align}
\label{eq3}
  \textbf{x'} = T(\textbf{x}, \omega_i); \quad \textbf{w} = H(\textbf{x}, \omega_i)
\end{align}
The density $\sigma$ and color $\textbf{c}$ can be then obtained by taking $\textbf{x'}$, $\textbf{w}$ along with direction $\textbf{d}$ and latent appearance code $\psi_i$ as input into template NeRF $F$:
\begin{align}
\label{eq4}
  (\sigma, \textbf{c}) = F(\textbf{x'}, \textbf{w}, \textbf{d}, \psi_i)
\end{align}
HyperNeRF can model time-varying shapes even with topological changes and can render different expressions of face avatar by using setting specific ambient coordinates in hyper-space. However, it is far from fine-grained control and fails to achieve per attribute control. Inspired by HyperNeRF, CoNeRF introduces regressors $A$ and $M$ to regress the attribute $\alpha$ and the corresponding mask $\textbf{m}$. The attribute $\alpha$ is generated from latent deformation code $\omega_i$ and then is taken as input into ambient slicing surface $H$ to generate a coordinate in ambient space $\textbf{w}$:
\begin{align}
\label{eq5}
  \alpha = A(\omega_i); \quad \textbf{w} = H(\textbf{x}, \alpha)
\end{align}
The corresponding mask map $\textbf{m}$ is generated using the warped coordinate $\textbf{x'}$ and the ambient space $\textbf{w}$ and  then is used to mask out $\textbf{w}$:
\begin{align}
\label{eq6}
  \textbf{m} = M(\textbf{x'}, \textbf{w}); \quad \textbf{w'} = \textbf{w} \odot \textbf{m}
\end{align}
The following density and color generation is the same as HyperNeRF. CoNeRF maps attribute $\alpha$ to its corresponding area through mask \textbf{m} so as to achieve the control over the corresponding area when rendering from novel views by assigning attribute $\alpha$ with different values. However, it requires manual labeling of attribute $\alpha$ and the corresponding mask $\textbf{m}$ which is labor-intensive and can only perform simple control over each area since only one attribute is related to each mask. CoNeRF can also lead to movement in another area when controlling one area since its network architecture does not achieve complete decoupling between attributes.

\begin{figure*}[htp]
  \centering
  \includegraphics[width=0.80\textwidth]{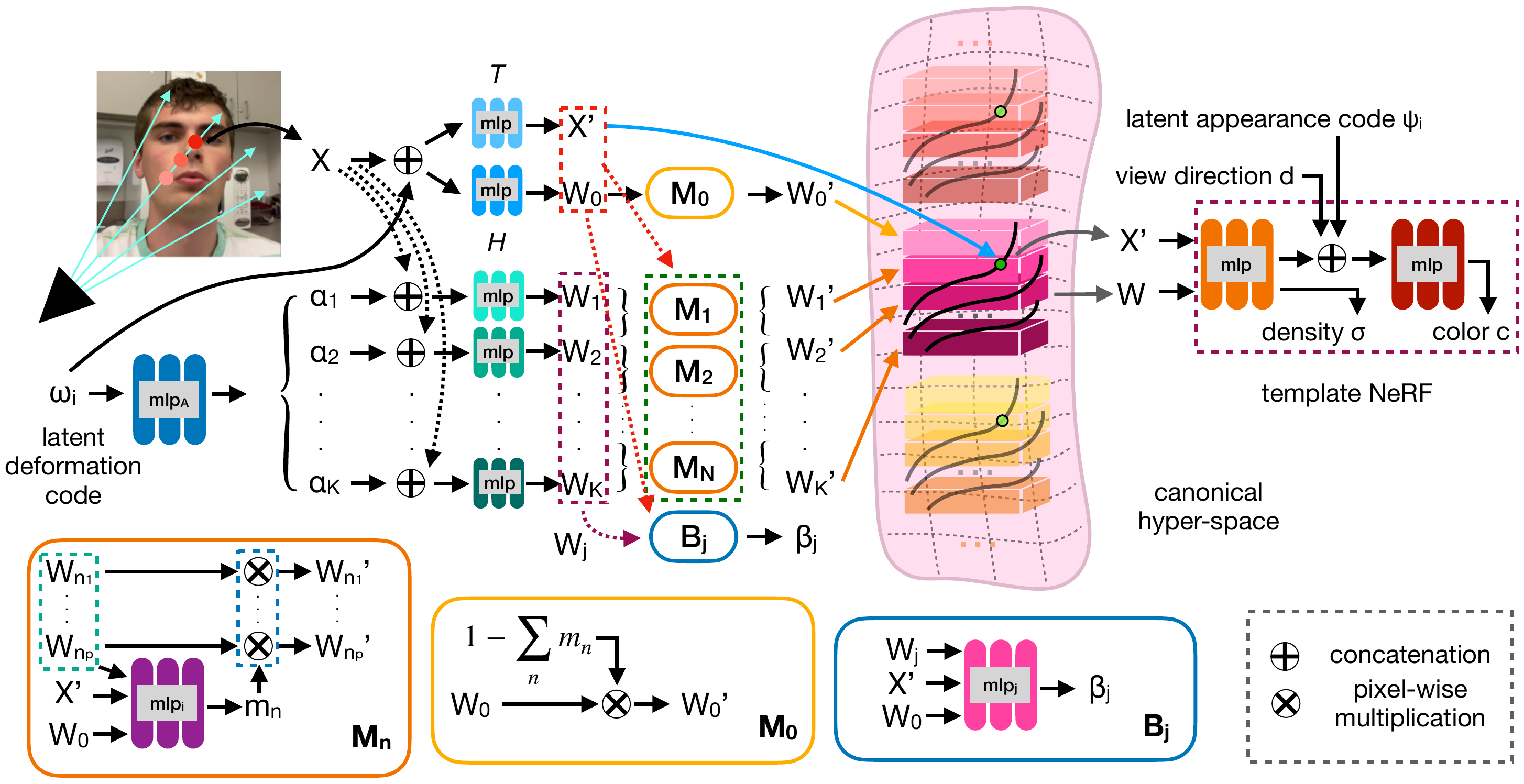}
  \caption{CoNFies architecture. $\alpha_i$ represents AU attribute learned from the latent code. $\beta_i$ is uncertainty and $m_i$ is mask.}
  \label{net}
\end{figure*}

\subsubsection{CoNFies. }
We propose CoNFies based on CoNeRF that can achieve more complex and accurate control over attributes. The network architecture of CoNFies is shown as Fig.~\ref{net}. We learn $K$ attributes $\alpha_{1\cdots K}$ from latent deformation code $\omega_i$ using attribute network $A$. Different from CoNeRF, we use $tanh$ as activation function in the last layer of $A$ to learn attributes whose range is $[-1, 1]$. We adopt the hyper-space $W$ as proposed in HyperNeRF while our hyper-space $W$ consists of $K+1$ ($K$ attributes and $1$ remaining part) components, which are attribute-specific. First, $K$ original hyper-space components are generated using $X$ and corresponding attribute $\alpha$ as following:
\begin{align}
\label{eq7}
  \textbf{w}_i = H_i(\textbf{x}, \alpha_i); \quad i=1\cdots K
\end{align}
We also generate one hyper-space $\textbf{w}_0$ for remaining avatar part and generate a warped coordinate \textbf{x'} through deformation field $T$:
\begin{align}
\label{eq8}
  \textbf{w}_0 = H_0(\textbf{x}, \omega_i); \quad \textbf{x'} = T(\textbf{x}, \omega_i)
\end{align}
After obtaining the original hyper-space $\textbf{w}_{0 \cdots K}$, we then learn $N$ masks according to a pre-defined correspondence between attributes and masks (many-to-one):
\begin{align}
\label{eq9}
  \textbf{m}_n = M_n(\textbf{x'}, \textbf{w}_0, \textbf{w}_{n_1} \cdots \textbf{w}_{n_p}); \quad n=1\cdots N
\end{align}
where $\textbf{w}_{n_1} \cdots \textbf{w}_{n_p}$ are the hyper-space generated from corresponding attributes $\alpha_{n_1} \cdots \alpha_{n_p}$ that are related to $\textbf{m}_n$. The mask $\textbf{m}_0$ for hyper-space $\textbf{w}_0$ is $1-\sum_n\textbf{m}_n$. Final hyper-space $\textbf{w}_{0 \cdots K}'$ are obtained by masking the original hyper-space using corresponding masks:
\begin{align}
\label{eq10}
  \textbf{w}_i' =\textbf{w}_i \odot \textbf{m}_j; \quad i=1\cdots K; \quad j=1\cdots N
\end{align}
where $\textbf{m}_j$ is the mask which $\alpha_i$ is related to. The whole hyper-space $W$ is obtained by concatenating $\textbf{w}_{0 \cdots K}'$ and the final density $\sigma$ and color $\textbf{c}$ are obtained using (\ref{eq4}), which is the same as template NeRF. We also render the mask field into image space using an analogous volume rendering process:

\begin{align}
\label{eq11}
  M(\textbf{r}|\theta, \beta_c) = \int_{t_n}^{t_f}T(t)\sigma(\textbf{r}(t))\textbf{m}(\textbf{r}(t),d)dt
\end{align}

It is also worth noting that we learn an uncertainty $\beta_i$ for each attribute $\alpha_i$ as follows to help reduce the potential noises after AUs filtering during training.
\begin{align}
\label{eq12}
  \beta_i = B_i(\textbf{x'}, \textbf{w}_0, \textbf{w}_i); \quad i=1\cdots K
\end{align}

\subsubsection{Training Losses. }
Given a training set collection of $C$ images, the losses of our method consist of two parts, reconstruction losses $L_{rec}$ and control losses $L_{ctrl}$, which are similar to ~\cite{kania2022conerf}:
\begin{align}
\label{eq13}
  \arg \min_{\theta, \{\mu_c\}} L_{rec}(\theta, \{\mu_c\}) + L_{ctrl}(\theta, \{\mu_c\})
\end{align}
where $\theta$ is network parameters and $\mu_c$ represents the latent code (deformation/appearance) of image $c$. Reconstruction losses $L_{rec}$ have two parts ($L_{recon}$ and $L_{reg}$). One is the primary reconstruction loss, which aims to reconstruct input observations $\{C_c\}$ as follow (gt = ground truth):
\begin{align}
\label{eq14}
  L_{recon} = \sum_{\textbf{r}\in R}\|C(\textbf{r}|\theta, \beta_c)-C^{gt}(\textbf{r})\|_2^2
\end{align}
The other one a Gaussian prior on the latent codes $\{\mu_c\}$ as proposed in ~\cite{park2019deepsdf}:
\begin{align}
\label{eq15}
  L_{reg} = \sum_c\|\mu_c\|_2^2
\end{align}

Control losses $L_{ctrl}$ also have two parts: attribute mask loss $L_{mask}$ and attribute value loss $L_{attr}$, as proposed in ~\cite{kania2022conerf}. For attribute mask loss, we first project 3D volumetric neural mask field $\textbf{m}$ into 2D mask image using (\ref{eq11}) and the attribute mask loss can be written as:
\begin{align}
\label{eq16}
  L_{mask} = \sum_{\textbf{r},a}\delta_{c,a}CE(M(\textbf{r}|\theta, \beta_c), M_{c,a}^{gt}(\textbf{r}))
\end{align}
where $CE(\cdot,\cdot)$ represents cross entropy and $M_{c,a}^{gt}(\textbf{r})$ is $a$-th attribute in the $c$-th image. $\delta_{c,a}$ denotes an indicator, where $\delta_{c,a}=1$ means attribute $a$  for image $c$, which $\textbf{r}$ is belong to, is provided, otherwise $\delta_{c,a}=0$. We also stop gradients in (\ref{eq16}) w.r.t $\sigma$ and employ focal loss~\cite{lin2017focal} in place of the standard cross entropy loss as in ~\cite{kania2022conerf}. For attribute value loss, we employ the AUs after filtering as ground-truth and $\beta_{c,a}$ learned from (\ref{eq12}) to further reduce noises in AUs:
\begin{align}
\label{eq17}
  L_{attr} = \sum_c\sum_a\delta_{c,a}\frac{|\alpha_{c,a}-\alpha_{c,a}^{gt}|^2}{2\beta_{c,a}^2}+\frac{(\log \beta_{c,a})^2}{2}
\end{align}
where larger $\beta_{c,a}$ values attenuate the importance of learned $\alpha_{c,a}$ and the second term precludes the trivial minimum at $\beta_{c,a}=\infty$. Hence the network can better learn to adjust $\alpha_{c,a}$ and reduce the negative effect of noises in AUs. The final loss is:
\begin{align}
\label{eq18}
  L = L_{recon} + w_{reg}L_{reg} + w_{mask}L_{mask} + w_{attr}L_{attr}
\end{align}
where $w_{reg}$, $w_{mask}$ and $w_{attr}$ are weighting coefficients.

%%%%%%%%%%%%%%%%%%%%%%%%%%%%%%%%%%%%%%%%%%%%%%%%%%%%%%%%%%%%%%%%%%%%%%%%%%%%%%%%

\section{EXPERIMENTS}
\subsection{Implementation details}
Our method is based on the JAX~\cite{jax2018github} implementation of CoNeRF~\cite{kania2022conerf}. Attribute network $A$ has six layers, each of which is a 32 neuron multi-layer perceptron (MLP) and has a skip connection at the fifth layer following~\cite{park2021nerfies}~\cite{park2021hypernerf}~\cite{kania2022conerf}. Deformation field
$T$ and ambient slicing surface $H_i$ have the same architecture of those in ~\cite{park2021hypernerf}~\cite{kania2022conerf}. Mask network $M_i$ and uncertainty network $B_i$ have the same structure, which is a four-layer MLP with 128 neuron per layer and followed by an additional 64 neuron layer with a skip connection as in ~\cite{kania2022conerf}. The template NeRF is the same as original NeRF~\cite{mildenhall2020nerf} but with a different input dimension size.
In our case, the number of attributes $K$ is 17, which is the number of AUs and the number of mask $N$ is 3 as shown in Fig.~\ref{anno}. We resize all the input images to $480\times 270$ and train our NeRF model for 250k iterations with 128 samples per ray and a batch size of 512 rays. We use Adam~\cite{kingma2014adam} with initial learning rate $lr=1e-4$ and set $w_{reg}=1e-4$, $w_{mask}=1e-2$ and $w_{attr}=0.1$ for all experiments. We introduce exponentially decaying on $lr$ and $w_{attr}$, which decay to $1e-5$ and $0$, respectively. Exponentially decaying on $w_{attr}$ can help further reduce the noises introduced by AU detection. We train our model on a NVIDIA A100 GPU with 80G memory and the whole process takes around 26 hours

\begin{figure}[t]
	\centering
	\subfigure[eye control]{
		\begin{minipage}{0.4\textwidth} 
            \includegraphics[width=\textwidth]{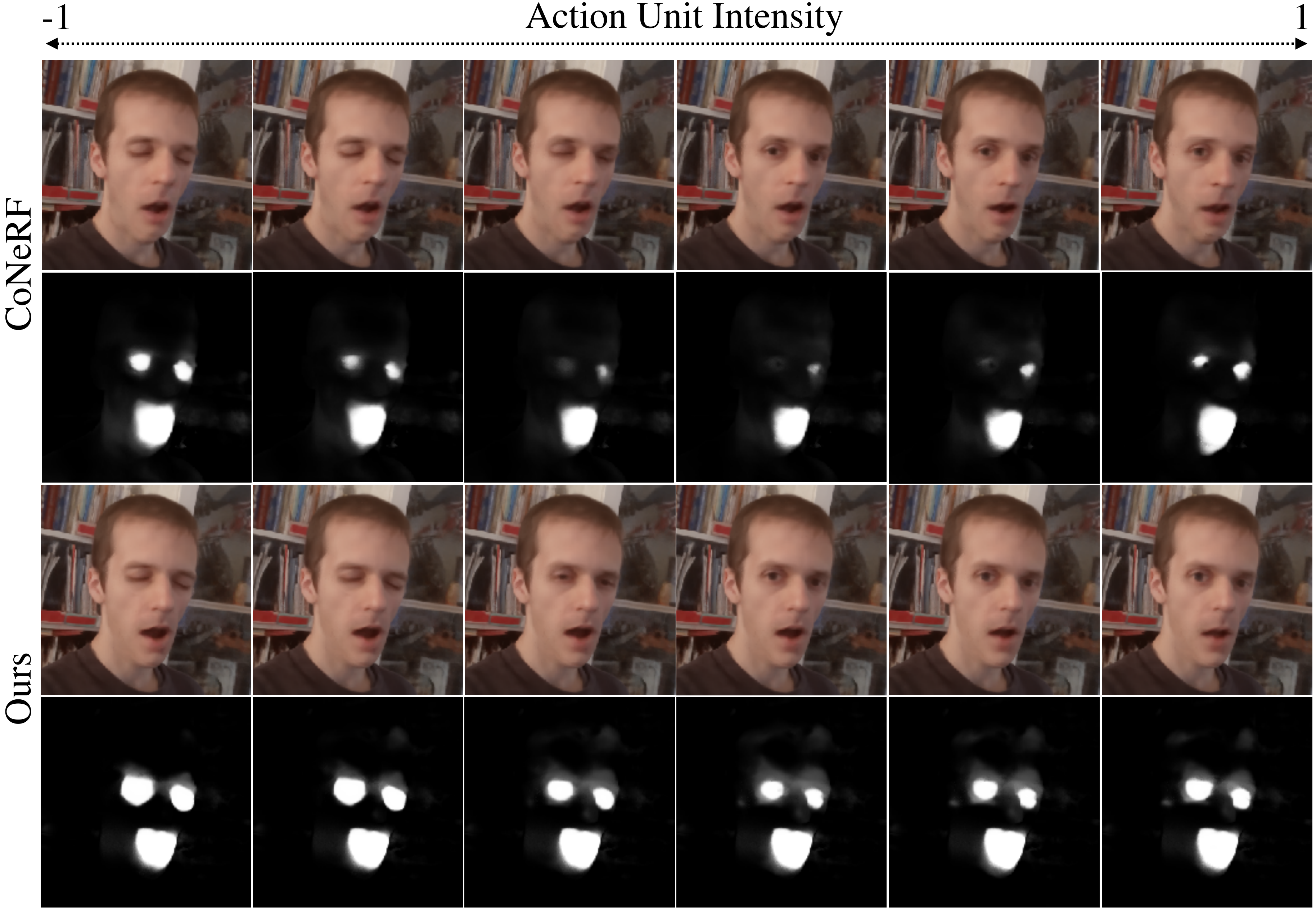} \\
		\end{minipage}
	}
	\subfigure[mouth control]{
		\begin{minipage}{0.4\textwidth}
			\includegraphics[width=\textwidth]{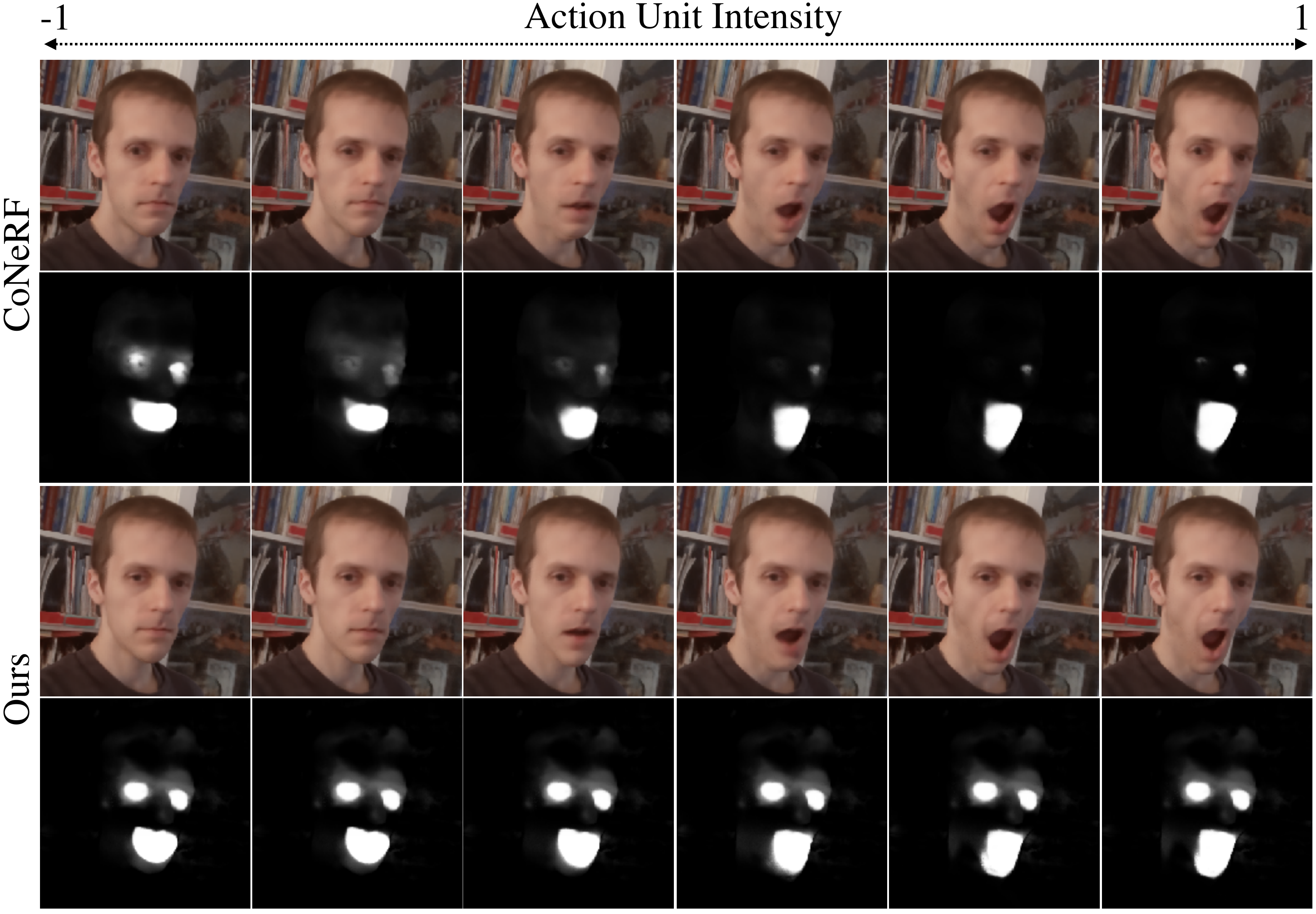} \\
			
		\end{minipage}
	}
	\caption{Controlling results of CoNFies and CoNeRF. Our CoNFies can perform a better control over one attribute without affecting other attributes.} 
	\label{comp}
\end{figure}

\begin{figure}[h!]
  \centering
  \includegraphics[width=0.4\textwidth]{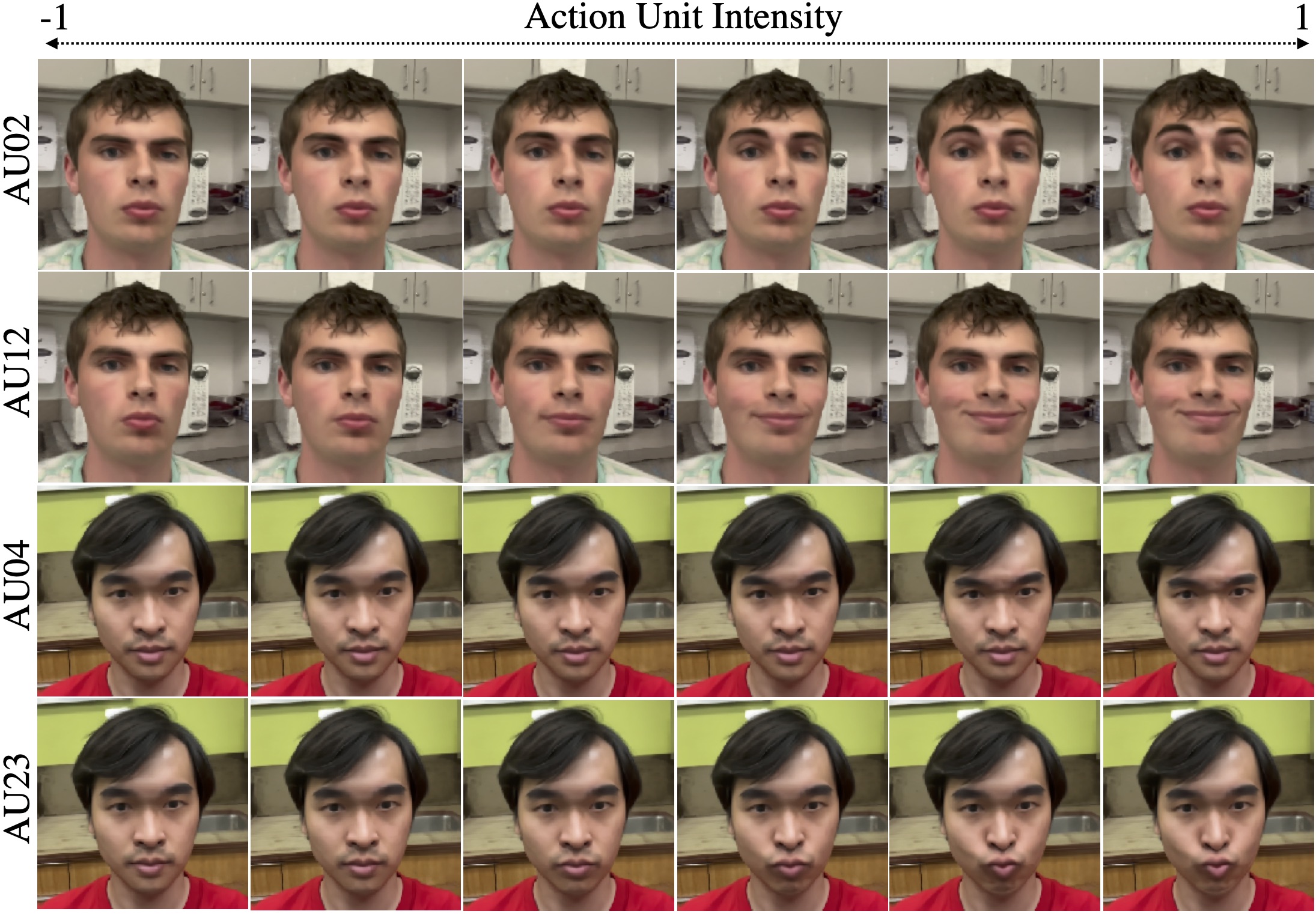}
  \caption{Control using single AU. AU02 is outer brow raiser. AU04 is brow lowerer. AU12 is lip corner puller. AU23 is lip tightener.}
  \label{perau}
\vspace{-0.1cm}
\end{figure}

\subsection{Dataset}
We used both the real dataset provided in \cite{kania2022conerf} and we collected our own video sequences using a smartphone. In our data collection each of the sequences was captured with an Apple iPhone 13 Pro using 120 fps slo-mo mode and is about 2 minutes long. In each video sequence, the person performs different facial expressions related to a single AU one by one first and then performs arbitrary facial expressions related to multiple AUs. Each video is extracted to frames with 120 fps and we perform OpenFace ~\cite{baltrusaitis2018openface} and smoothing as mentioned above. Note that OpenFace can only provide 17 AU intensities and we use these in the following experiments. More AU intensities may be obtained from other methods, which is not the focus of our paper. We then undersample the sequences to give approximately 750 frames per capture. The frames along with the AUs and attribute masks generated automatically form the datasets we use in our experiments. 

\subsection{Decoupling Mask}
We compare our method with CoNeRF using their dataset with manual attribute values and mask area labels to show the effectiveness of our decoupling mask structure. We control the eyes and mouth separately using attribute values and show the rendering image results along with the masks in Fig.~\ref{comp}. We can see from the mask results that when controlling one attribute, the masks of the other attribute are not affected in our method. But one attribute can affect the others in CoNeRF, which lead to unexpected movement and artifacts in the rendering results.

\subsection{Attribute Control}
Our CoNFies can achieve attribute control using different AUs. We first show the controlling result using single AU (AU 02, 04, 12, 23) on two sequences as in Fig.~\ref{perau}. 
%To quantitatively evaluate our method, we also run CoNeRF on the same sequences and only use extreme AU values detected by OpenFace as labels. We then perform control on each AU and run OpenFace to detect the truth AUs on the rendering results of our method and CoNeRF. We calculate ICC as shown below, which indicates our methods performs much better than CoNeRF.
We also conducted quantitative evaluation to compare our method and CoNeRF. 

\begin{table}[t!]
\caption{Intraclass Correlation (ICC) comparison between CoNeRF and our method.}
\label{tab:icc}
\begin{center}
\begin{tabular}{|c|cc|c|c|cc|}
\cline{1-3} \cline{5-7}
AU & CoNeRF & Ours &  & AU   & CoNeRF & Ours \\ \cline{1-3} \cline{5-7} 
01  & 0.52   & 0.86 &  & 14   & 0.17   & 0.77 \\
02  & 0.54   & 0.73 &  & 15   & -0.15  & 0.81 \\
04  & 0.23   & 0.91 &  & 17   & 0.31   & 0.88 \\
05  & -0.11  & 0.40 &  & 20   & 0.03   & 0.63 \\
06  & 0.00   & 0.00 &  & 23   & 0.49   & 0.82 \\
07  & -0.44  & 0.73 &  & 25   & 0.36   & 0.90 \\
09  & 0.55   & 0.74 &  & 26   & 0.08   & 0.93 \\
10 & 0.00   & 0.00 &  & 45   & 0.21   & 0.83 \\ \cline{5-7} 
12 & 0.00   & 0.87 &  & mean & 0.16   & 0.69 \\ \cline{1-3} \cline{5-7} 
%\hline
        % 1 & 0.52 & 0.86 \\
        % 2 & 0.54 & 0.73 \\
        % 4 & 0.23 & 0.91 \\
        % 5 & -0.11 & 0.40 \\
        % 6 & 0.00 & 0.00 \\
        % 7 & -0.44 & 0.73 \\
        % 9 & 0.55 & 0.74 \\
        % 10 & 0.00 & 0.00 \\
        % 12 & 0.00 & 0.87 \\
        % 14 & 0.17 & 0.77 \\
        % 15 & -0.15 & 0.81 \\
        % 17 & 0.31 & 0.88 \\
        % 20 & 0.03 & 0.63 \\
        % 23 & 0.49 & 0.82 \\
        % 25 & 0.36 & 0.90 \\
        % 26 & 0.08 & 0.93 \\
        % 45 & 0.21 & 0.83 \\ \hline
        % mean & 0.16 & 0.69 \\
        % au & conerf & ours & ~ & au & conerf & ours \\ \hline
        % 01 & 0.52 & 0.86 & ~ & 14 & 0.17 & 0.77 \\
        % 02 & 0.54 & 0.73 & ~ & 15 & -0.15 & 0.81 \\
        % 04 & 0.23 & 0.91 & ~ & 17 & 0.31 & 0.88 \\
        % 05 & -0.11 & 0.40 & ~ & 20 & 0.03 & 0.63 \\
        % 06 & 0.00 & 0.00 & ~ & 23 & 0.49 & 0.82 \\
        % 07 & -0.44 & 0.73 & ~ & 25 & 0.36 & 0.90 \\
        % 09 & 0.55 & 0.74 & ~ & 26 & 0.08 & 0.93 \\
        % 10 & 0.00 & 0.00 & ~ & 45 & 0.21 & 0.83 \\
        % 12 & 0.00 & 0.87 & ~ & mean & 0.16 & 0.69 \\
%\hline
\end{tabular}
\end{center}
% \vspace{-0.7cm}
\end{table}

\begin{figure}[h!]
  \centering
  \includegraphics[width=0.35\textwidth]{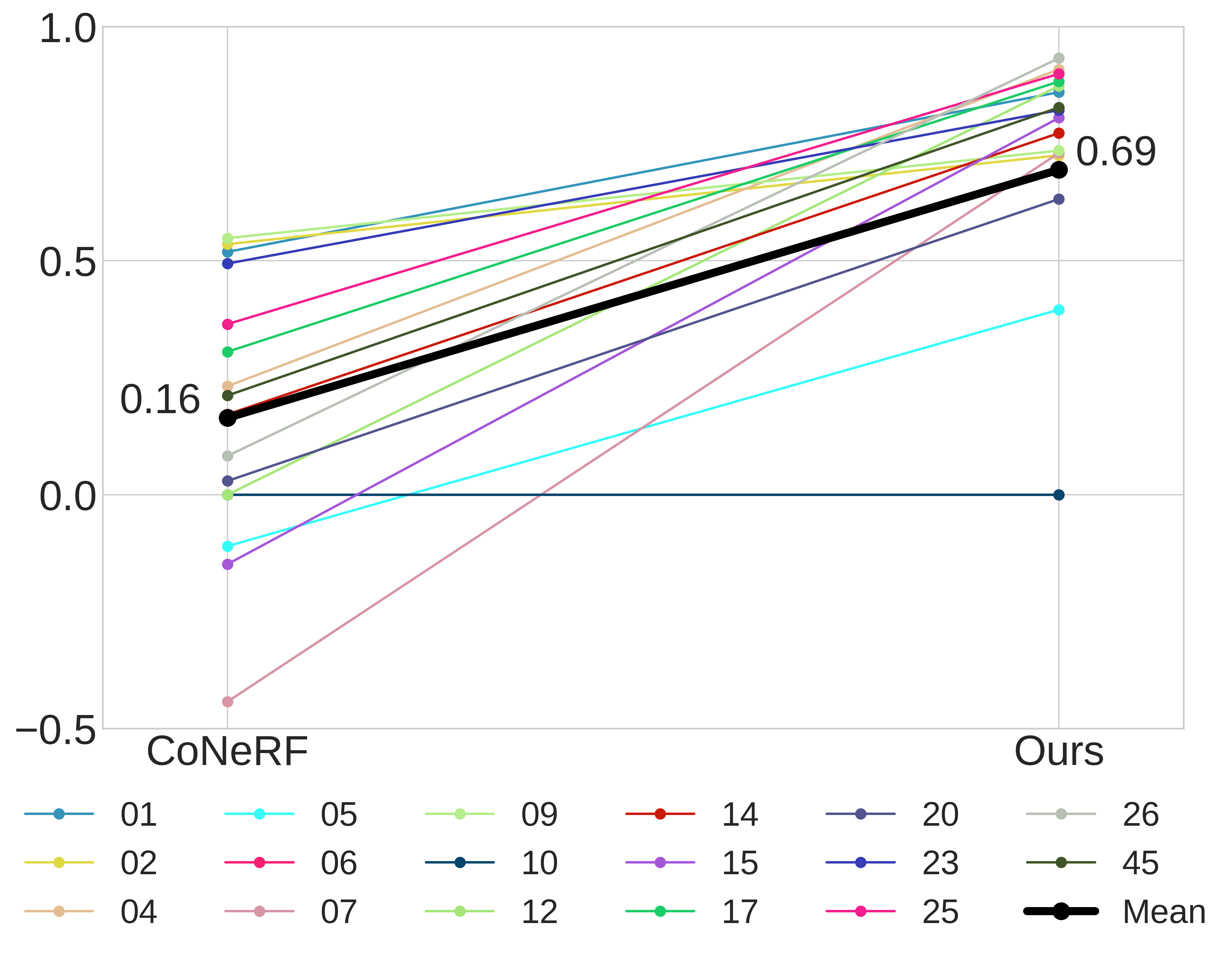}
  \caption{Intraclass Correlation (ICC) comparison between CoNeRF and our method.}
  \label{fig:icc}
  \vspace{-0.5cm}
\end{figure}

\begin{figure}[h!]
  \centering
  \includegraphics[width=0.25\textwidth]{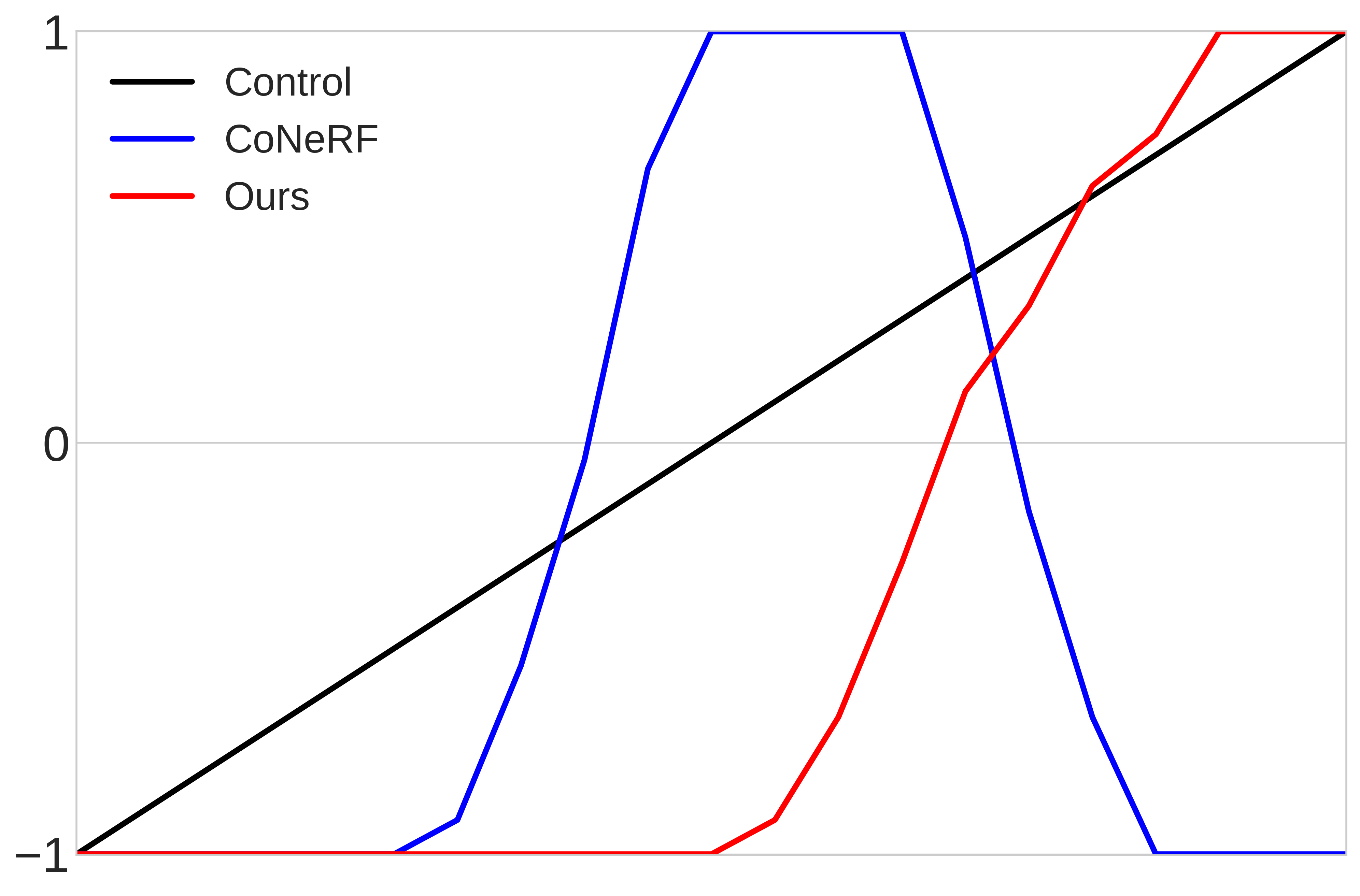}
  \caption{Comparison of AU01 intensity transition between control values and synthesized images.}
  \label{fig:icc_trans}
\end{figure}

In this experiment, AU intensities were obtained from synthesis images generated by our method and CoNeRF, and compare them using Intraclass Correlation (ICC). 
OpenFace was used to obtain AU intensities from synthesis images. Only images with extreme AU values (near -1 or 1) were manually selected to train CoNeRF while our method automatically obtained images to train the model. Table~\ref{tab:icc} and Fig.~\ref{fig:icc} show that our method outperforms CoNeRF. Fig.~\ref{fig:icc_trans} compares AU01 intensity transition between control values and synthesized images. The AU intensity transition of our method is much closer to the control than CoNeRF. This result also indicates that our method can handle subtle AU changes better than CoNeRF.

\begin{figure}[t]
  \centering
  \includegraphics[width=0.40\textwidth]{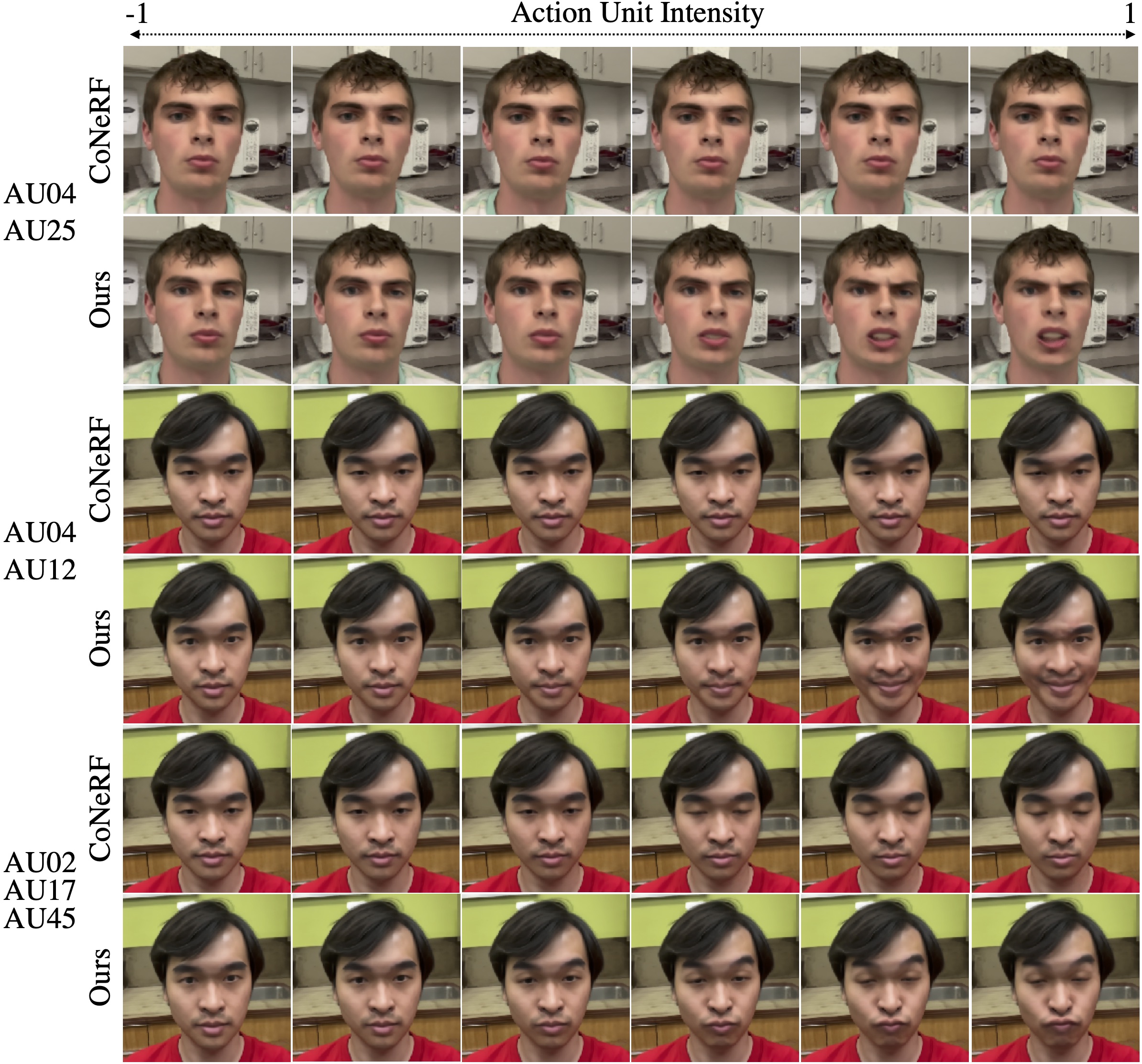}
  \caption{Control using multiple AUs on different regions.  AU02 is outer brow raiser. AU04 is brow lowerer. AU12 is lip corner puller. AU17 is chin raiser. AU25 is lips part. AU45 is blink.}
  \label{multiau}
\end{figure}

\begin{figure}[h!]
  \centering
  \includegraphics[width=0.40\textwidth]{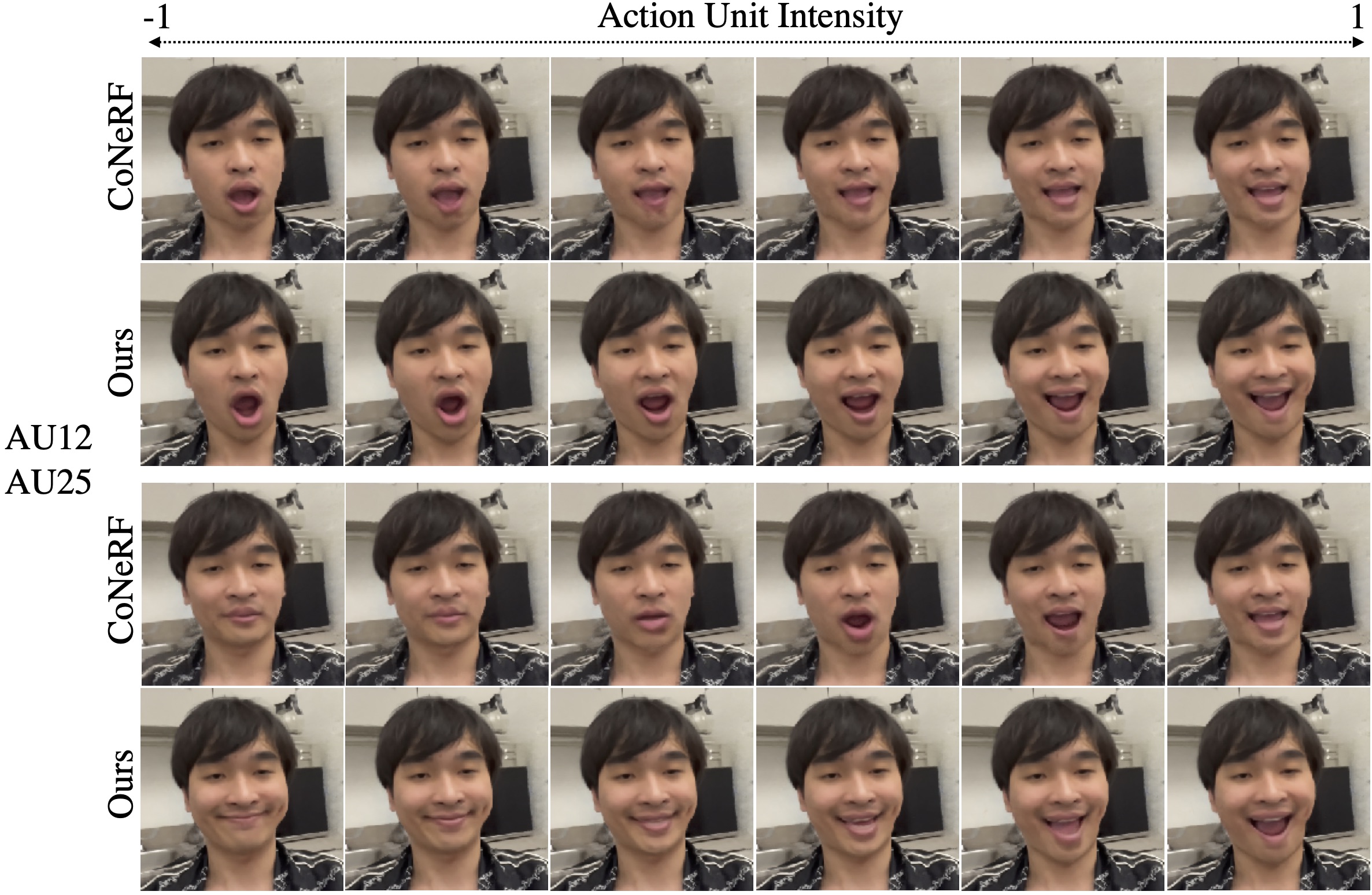}
  \caption{Control using multiple AUs on the same region. AU12 is lip corner puller. AU25 is lips part.}
  \label{multiau1}
\end{figure}

Our method can perform complex control using multiple AUs simultaneously as shown in Fig.~\ref{multiau} and Fig.~\ref{multiau1}. From Fig.~\ref{multiau}, we can see that our method can perform combined AUs control over different regions, e.g. eyebrow and mouth. 

We show that our method can even perform more complicated control over the same region. As shown in Fig.~\ref{multiau1}, we can control the smile while keeping mouth open or control mouth open while keeping the smile. However, CoNeRF cannot render good results under such complex scenarios. More visualization results can be found in the supplementary material.

% \begin{figure}[htp]
%   \centering
%   \includegraphics[width=0.45\textwidth]{FG2023_Latex_template/imgs/per_au_txt.pdf}
%   \caption{Control using single AU}
%   \label{perau}
% \end{figure}

\subsection{Novel View Synthesis}

\begin{figure}[t]
  \centering
  \includegraphics[width=0.40\textwidth]{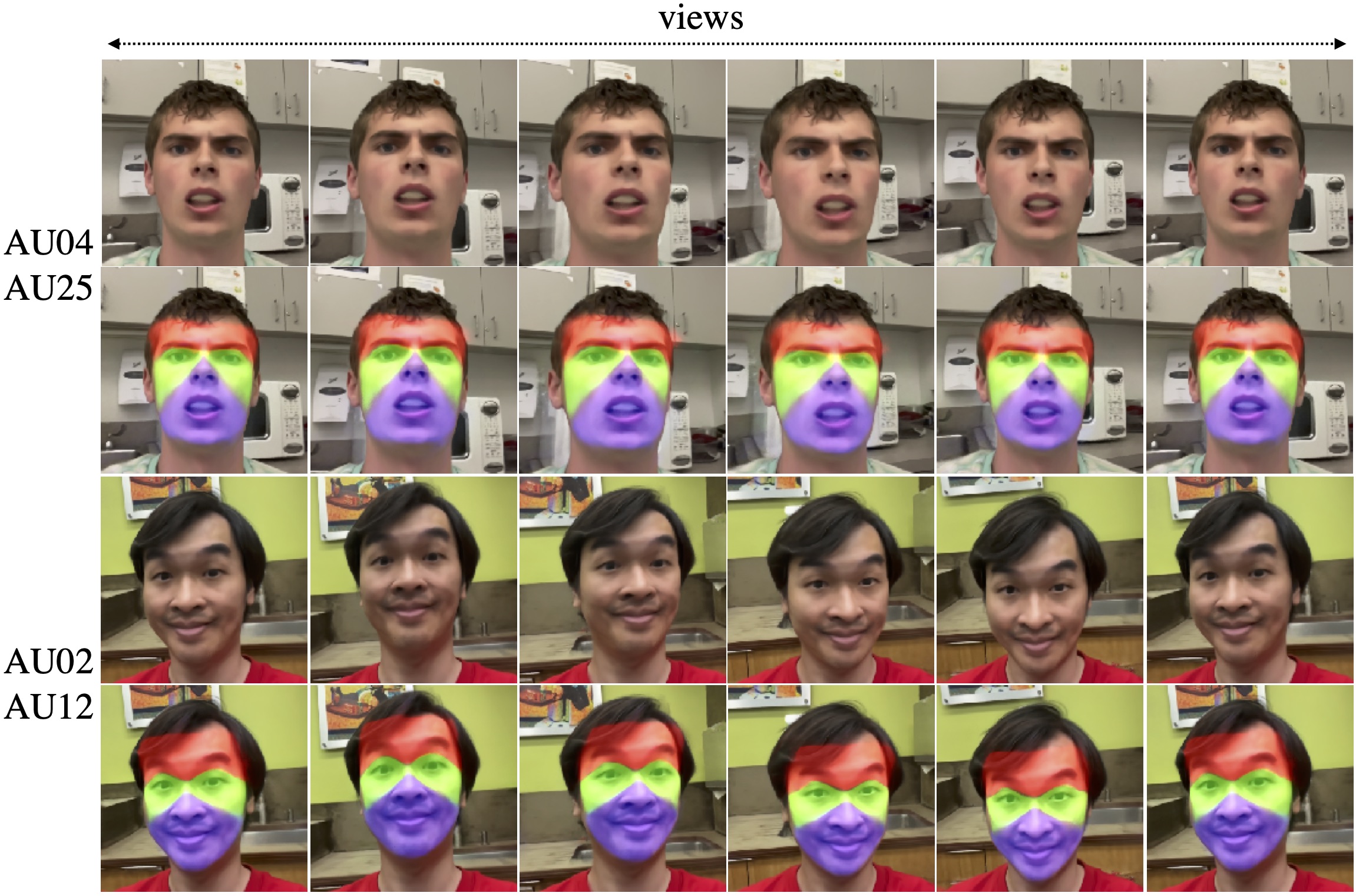}
  \caption{Novel view synthesis under fixed AU setting. AU02 is outer brow raiser. AU04 is brow lowerer. AU12 is lip corner puller. AU25 is lips part.}
  \label{fixau}
  \vspace{-0.5cm}
\end{figure}

\begin{figure}[h!]
  \centering
  \includegraphics[width=0.37\textwidth]{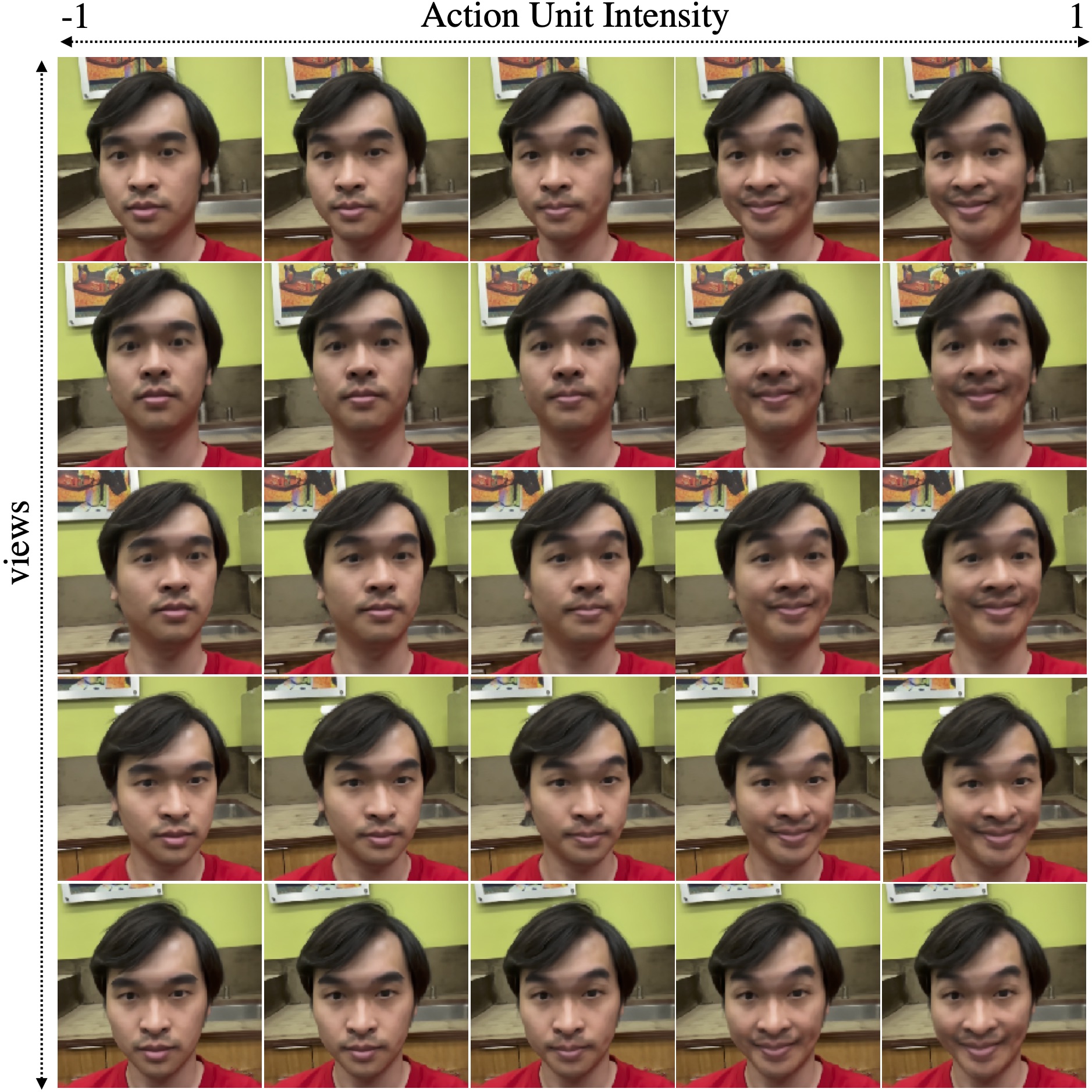}
  \caption{Rendering results under different AU (AU02 and AU12)  intensities and views}
  \label{attrview}
\end{figure}

As a NeRF-based method, we can synthesize novel views with single or multiple AUs control. We show the novel view synthesis results along with corresponding masks in Fig.~\ref{fixau}, where we keep the facial expression constant. To further show the power of our method, we show the rendering result under different views and control the AU values (AU02 and AU12) simultaneously in Fig.~\ref{attrview}.

We also evaluate the rendering quality of our method on a frame interpolation task as proposed in ~\cite{kania2022conerf} using the dataset it released. 
We interpolate every other frame and do not perform any attribute control. We use Peak Signal-to-Noise Ratio (PSNR), Multi-scale Structural Similarity (MS-SSIM)~\cite{wang2003multiscale} and Learned Perceptual Image Patch Similarity (LPIPS)~\cite{zhang2018unreasonable} to quantitatively evaluate our method compared with NeRF~\cite{mildenhall2020nerf}, NeRF + Latent, Nerfies~\cite{park2021nerfies}, HyperNeRF~\cite{park2021hypernerf}, CoNeRF-$M$ and CoNeRF~\cite{kania2022conerf}, which is consistent with ~\cite{kania2022conerf}. The result in shown in Table~\ref{table2}, where we can see that our method can achieve comparable performance in the rendering quality from a novel view.

\begin{table}
\caption{Quantitative results}
\begin{center}
\begin{tabular}{|c||c|c|c|}
\hline
Method & PSNR$\uparrow$ & MS-SSIM$\uparrow$ & LPIPS$\downarrow$\\
\hline
NeRF & 28.795 & 0.951 & 0.210\\
NeRF + Latent & 32.653 & 0.981 & 0.182\\
NeRFies & 32.274 & 0.981 & 0.180\\
HyperNeRF & 32.520 & 0.981 & 0.169\\
CoNeRF-$M$ & 32.061 & 0.979 & 0.167\\
CoNeRF & 32.342 & 0.981 & 0.168\\
\hline
Ours & 32.356 & 0.982 & 0.166\\
\hline
\end{tabular}
\end{center}
\label{table2}
% \vspace{-0.7cm}
\end{table}

\begin{figure}[h!]
  \centering
  \includegraphics[width=0.4\textwidth]{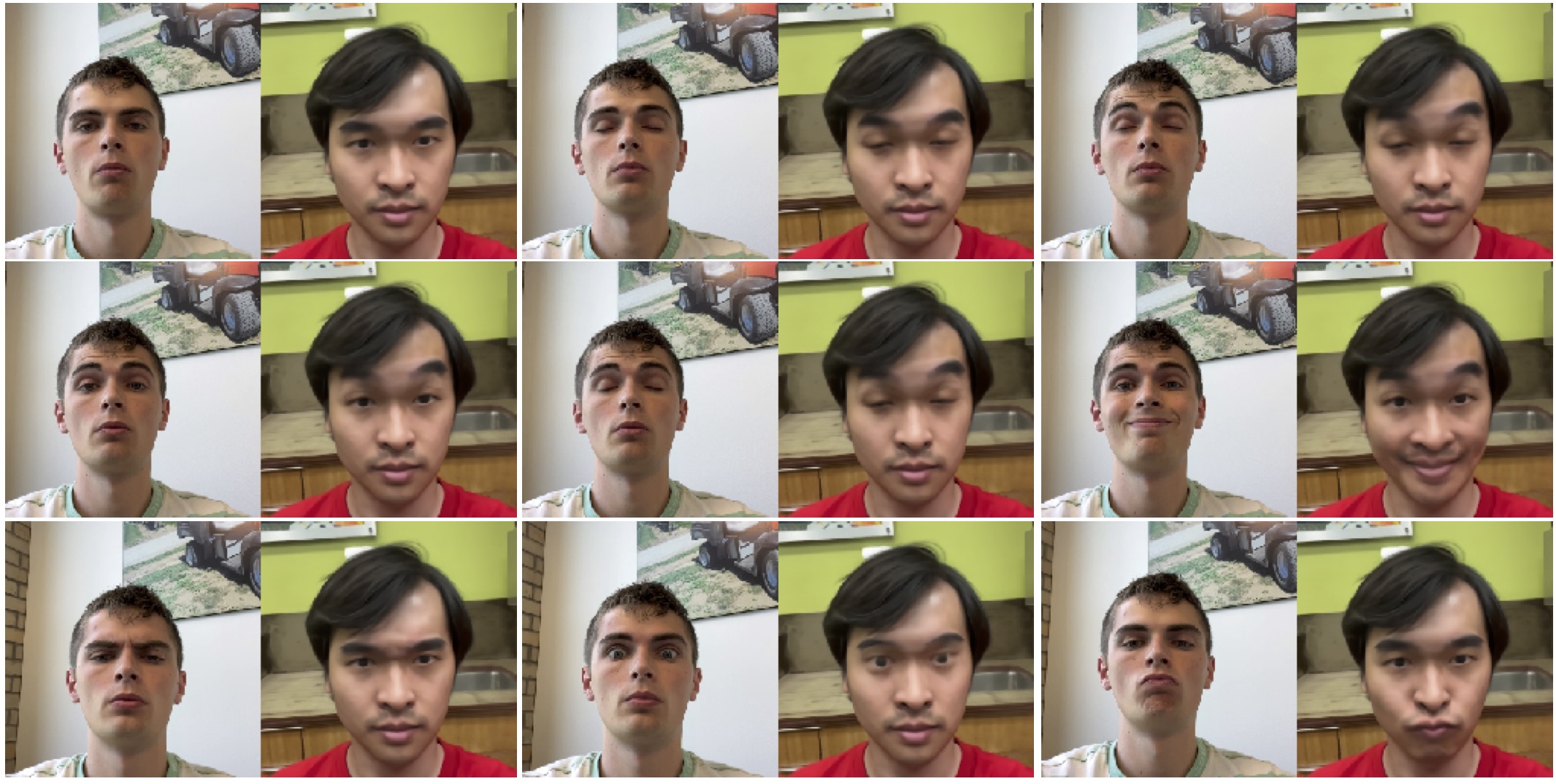}
  \caption{Facial expression transfer using reference sequence.}
  \label{anim}
  \vspace{-0.5cm}
\end{figure}

\subsection{Facial Expression Transfer}
In addition to adjusting AU values manually to control the avatar's expression, it is possible to copy them from another person's face. In this case, first we detect the AU intensities from the source person's face and use them to re-synthesize the same expression on the avatar. We show the rendering results using another face sequence to control the trained sequence in Fig.~\ref{anim}.

\section{CONCLUSIONS}
We have proposed an automated approach for controllable neural face avatars. Once a 2D video is recorded using slo-mo mode, our network is automatically trained by utilizing AU intensities and facial landmarks. We have introduced the decoupling mask structure so that the different semantic regions do not affect each other and each region have multiple control variables. We have shown that our approach outperforms CoNeRF in terms of fine-grained AU control. Moreover, our approach has a capability to control multiple AUs and novel views simultaneously.

\section{ACKNOWLEDGMENTS}
This research was supported by Fujitsu. We thank Joel Julin from University of Pittsburgh for helping with data collection and comments that greatly improved the manuscript. We thank Xuxin Cheng from Carnegie Mellon University who provided the data acquisition equipment. We  would also like to show our gratitude to Nian-Hsuan Tsai from Carnegie Mellon University who helped build the OpenFace system.

% The authors gratefully acknowledge the contribution of reviewers' comments, etc. (if desired). Put sponsor acknowledgments in the unnumbered footnote on the first page.

%%%%%%%%%%%%%%%%%%%%%%%%%%%%%%%%%%%%%%%%%%%%%%%%%%%%%%%%%%%%%%%%%%%%%%%%%%%%%%%%

% References are important to the reader; therefore, each citation must be complete and correct. If at all possible, references should be commonly available publications.

{\small
\bibliographystyle{ieee}
\bibliography{egbib}
}

\end{document}